\documentclass{article}
\usepackage{graphicx}

\usepackage{nac_preprint}
\usepackage[utf8]{inputenc} 
\usepackage[T1]{fontenc}    
\usepackage{hyperref}       
\usepackage{url}            
\usepackage{booktabs}       
\usepackage{amsfonts}       
\usepackage{nicefrac}       
\usepackage{microtype}      

\usepackage{booktabs} 
\usepackage{subcaption}
\usepackage{amsmath}
\usepackage{amssymb}
\usepackage{cleveref}
\usepackage{mathtools}
\usepackage[toc,page]{appendix}
\usepackage{enumitem}
\usepackage{graphics}
\usepackage{graphicx}
\usepackage{xcolor, colortbl}
\usepackage[export]{adjustbox}
\usepackage{algorithm}
\usepackage{algorithmicx}
\usepackage{multicol}
\usepackage{algpseudocode}
\usepackage[inkscapelatex=false]{svg} 

\definecolor{LightCyan}{rgb}{0.88,1,1}
\definecolor{LightOrange}{rgb}{1,0.76,0.86}

\algnewcommand{\LineComment}[1]{\State \(//\) #1}
\algnewcommand{\RLineComment}[1]{\State \(\triangleright\) #1}
\newlength{\algrhswidth}
\usepackage{bm}
\usepackage{multirow}
 
\usepackage{wrapfig}

\usepackage{etoolbox}
\usepackage{tikz}
\usetikzlibrary{tikzmark}
\usetikzlibrary{calc}

\errorcontextlines\maxdimen

\newcommand{\ALGtikzmarkcolor}{black}
\newcommand{\ALGtikzmarkextraindent}{4pt}
\newcommand{\ALGtikzmarkverticaloffsetstart}{-.5ex}
\newcommand{\ALGtikzmarkverticaloffsetend}{-.5ex}
\makeatletter
\newcounter{ALG@tikzmark@tempcnta} 

\newcommand\ALG@tikzmark@start{%
    \global\let\ALG@tikzmark@last\ALG@tikzmark@starttext%
    \expandafter\edef\csname ALG@tikzmark@\theALG@nested\endcsname{\theALG@tikzmark@tempcnta}%
    \tikzmark{ALG@tikzmark@start@\csname ALG@tikzmark@\theALG@nested\endcsname}%
    \addtocounter{ALG@tikzmark@tempcnta}{1}%
}

\def\ALG@tikzmark@starttext{start}
\newcommand\ALG@tikzmark@end{%
    \ifx\ALG@tikzmark@last\ALG@tikzmark@starttext
    \else
        \tikzmark{ALG@tikzmark@end@\csname ALG@tikzmark@\theALG@nested\endcsname}%
        \tikz[overlay,remember picture] \draw[\ALGtikzmarkcolor] let \p{S}=($(pic cs:ALG@tikzmark@start@\csname ALG@tikzmark@\theALG@nested\endcsname)+(\ALGtikzmarkextraindent,\ALGtikzmarkverticaloffsetstart)$), \p{E}=($(pic cs:ALG@tikzmark@end@\csname ALG@tikzmark@\theALG@nested\endcsname)+(\ALGtikzmarkextraindent,\ALGtikzmarkverticaloffsetend)$) in (\x{S},\y{S})--(\x{S},\y{E});%
    \fi
    \gdef\ALG@tikzmark@last{end}%
}

\apptocmd{\ALG@beginblock}{\ALG@tikzmark@start}{}{\errmessage{failed to patch}}
\pretocmd{\ALG@endblock}{\ALG@tikzmark@end}{}{\errmessage{failed to patch}}
\makeatother

\title{Enhancing Eye Feature Estimation from Event Data Streams through Adaptive Inference State Space Modeling}

\author{%
Viet Dung Nguyen\\
 Rochester Institute of Technology \\ 
\texttt{vn1747@rit.edu}
\And
Mobina Ghorbaninejad\\
 Rochester Institute of Technology \\ 
\texttt{mg2587@.rit.edu}
\And
Chengyi Ma\\
 Rochester Institute of Technology \\ 
\texttt{cxm3593@rit.edu}
\And
Reynold Bailey\\
 Rochester Institute of Technology \\ 
\texttt{rjbvcs@rit.edu}
\And
Gabriel J. Diaz\\
 Rochester Institute of Technology \\ 
\texttt{Gabriel.Diaz@rit.edu}
\And
Alexander Fix\\
 Meta Reality Labs \\ 
\texttt{alexander.fix@meta.com}
\And
Ryan J. Suess\\
 Meta Reality Labs \\ 
\texttt{ryan.suess@gmail.com}
\And
Alexander Ororbia \\
Rochester Institute of Technology \\
\texttt{ago@cs.rit.edu}
}

\begin{document}

\setlength{\abovedisplayskip}{0.065cm}
\setlength{\belowdisplayskip}{0pt}

\maketitle

\begin{abstract}

Eye feature extraction from event-based data streams can be performed efficiently and with low energy consumption, offering great utility to real-world eye tracking pipelines. However, few eye feature extractors are designed to handle sudden changes in event density caused by the changes between gaze behaviors that vary in their kinematics, leading to degraded prediction performance. In this work, we address this problem by introducing the \emph{adaptive inference state space model} (AISSM), a novel architecture for feature extraction that is capable of dynamically adjusting the relative weight placed on current versus recent information. This relative weighting is determined via estimates of the signal-to-noise ratio and event density produced by a complementary \emph{dynamic confidence network}. Lastly, we craft and evaluate a novel learning technique that improves training efficiency. Experimental results demonstrate that the AISSM system outperforms state-of-the-art models for event-based eye feature extraction.

\end{abstract}

\section{Introduction}
\label{sec:intro}

Many methods for gaze estimation rely on the localization of eye features present in near-eye images~\cite{mackworth1958eye}. Example features include the pupil boundary~\cite{Kothari2022EllSegGen, Kothari2021EllSeg}, pupil centroid~\cite{Chen2025EventBasedEyeTrackingChallenge2025, Chen20233ETCBConvLSTM}, or iris texture~\cite{Pelz2017GazeTrackingIris, Shah2009IrisSegmentation, Daugman1993IrisSegmentation}. Eye image collection typically relies on frame-based cameras operating at fixed frame rates and with roughly fixed exposure durations that can lead to motion blur and tracking latency. Due to the need for high spatial and temporal resolution, frame-based methods require a significant power draw and high bandwidth~\cite{Lichtsteiner2008EventBasedCameraOriginal}. 
For example, a binocular eye tracker with a resolution of $320 \times 320$ pixels operating at $90$ frames per second must transmit over $18$ million pixels per second, many of which do not meaningfully contribute to gaze estimation, e.g. surrounding skin, introducing inefficiencies in downstream feature extraction.

Recent research has explored \emph{event sensors} as a means to overcome many of the limitations of frame-based cameras ~\cite{Gallego2022EventBasedSurvey, Chakravarthi2024EventCamInnoSurvey, Iddrisu2024EventBasedEyeSurvey}. Event sensors, also known as dynamic vision sensors (DVS), passively monitor the scene and asynchronously transmit pixel-level `events' only when the observed change in log-luminance exceeds a threshold. This per-pixel, parallel sensing enables extremely low latency and effective sampling rates as high as $10,000$ Hertz~\cite{Gallego2022EventBasedSurvey}. In eye-tracking, event sensors offer several advantages over traditional frame-based sensors: 
\begin{enumerate}[noitemsep, topsep=2pt]
    \item they operate at a higher frequency and lower latency, leading to better accuracy in tracking saccades, micro-saccades, and fast eye movements; 
    \item since an event sensor only detects brightness/contrast changes, algorithms can exploit this motion-selective capturing behavior, i.e., there are abundant quantities of events when the eye moves, and nearly no events when the eye is in fixation; 
    \item event sensors consume significantly less power than frame-based cameras, resulting in more accessible wearable devices and edge devices~\cite{Chakravarthi2024EventCamInnoSurvey}; and, 
    \item when properly tuned, triggered events will be spatially sparse and localized primarily around task-relevant portions of the eye in motion, such as the iris-pupil boundary, where changes in luminance are greatest.
\end{enumerate}

Despite the advantages of event-based cameras for eye tracking, several challenges remain at the frontier of research. A key challenge arises from the highly variable kinematics of eye movements. Notably, the eye exhibits brief periods of rapid, ballistic \emph{saccadic} movements that generate dense bursts of task-relevant events, yielding a high signal-to-noise ratio (SNR). These bursts are interspersed with periods of slower or lower-amplitude eye movements (e.g., smooth pursuit, vestibulo-ocular reflex, and fixational eye movements), which elicit comparatively few events and result in a lower SNR. One potential mitigation strategy is to increase the temporal window over which events are accumulated prior to image formation and feature extraction. However, this approach comes at the cost of increased system latency as well as a potential reduction in effective sampling rate. Replacing a convolutional neural network (CNN) with a recurrent neural network (RNN), as is done in some efforts~\cite{Lipton2015RNNSurvey, Chung2014GRU}, presents similar trade-offs. Beyond latency-related concerns, recurrence is computationally inefficient during periods of rapid eye movement when SNR is already high and it may introduce bias by incorporating recent temporal context when it is not necessary.

In this work, we test the hypothesis that relying only on a history-driven or recurrent model leads to inaccurate feature extraction in the presence of abrupt changes in the event density or SNR. We further propose and evaluate a novel Bayesian model designed to overcome these issues through an adaptive neural mechanism that dynamically weights the model's posterior and prior. This scheme -- the \emph{adaptive inference state space model (AISSM)} -- is inspired by the Kalman filter~\cite{Kalman1960Filter} and aims to leverage information during periods of high SNR, while incorporating historical information when SNR drops. Furthermore, we introduce a module called the \emph{dynamic confidence network}, which aids the AISSM in learning from dynamic behavior. To adapt the parameters of these neuronal components, we construct a model learning technique called \emph{long-horizon training} in order to improve the training efficiency and maintain the accuracy of the final model. Evaluation demonstrates that our approach outperforms other event-based feature tracking algorithms. Additional testing provides insight into the contributions of the proposed modules.

\section{Related Work}
\label{sec:related_work}

\subsection{Event Representations}
\label{sec:event_rep}

Instead of producing a dense pixel frame like conventional cameras do, a DVS generates a spatially sparse stream of events. Each event is denoted as a tuple $(x, y, p, t)$ where $x$ and $y$ are the pixel coordinates, $t$ is the event timestamp, and $p$ is the binary polarity of the event, indicating an increase ($1$) or decrease ($-1$) in light intensity at that pixel~\cite{Lichtsteiner2008EventBasedCameraOriginal}. 
Collectively, events form a 3D event stream across the sensor plane and over time; researchers have begun to explore ways to leverage this raw stream for feature extraction. \emph{EyeGraph}~\cite{Bandara2024EyeGraph} applies graph-based modeling to encode local spatio-temporal relationships and was shown to work well as a feature extraction method that was robust to the variability in eye movement kinematics in terms of SNR (albeit with a computational cost and latency that hindered deployment in real-time systems). Other approaches have adopted representations that `bin' events into frames to allow the frame-duration to increase in response to sparsity. For instance, a histogram-based approach~\cite{liu2018adaptive} aggregates events by counting their occurrences at each pixel location within a fixed temporal window and a \emph{time surface} records the timestamp of the most recent event at each pixel; this creates a map where brightness is indicative of motion recency~\cite{delbruck2008frame, Lagorce2017EventBasedTimeSurface}. The \emph{voxel-grid} is a compromise in which the $3$D spatio-temporal domain is quantized into voxels spanning both time and the $x$ and $y$ dimensions of the sensor plane~\cite{truong2025dual}. 
In our work, we adopt a simple binary event-frame representation, i.e., \emph{binarep}~\cite{Barchid2022BinaRepEventFrames}, which combines event polarity into a lightweight $2$D grid compatible with neural networks. Paired with our proposed temporal model, this approach extracts sufficient spatial structure in order to achieve state-of-the-art performance without added representation complexity.

\subsection{Model Architectures}
\label{sec:model_arch}

Beyond considering different event representations, researchers have explored artificial neural network (ANN) architecture designs that mitigate the effects of variable SNR on feature extraction performance. Due to the lack of established methodologies for directly processing raw spatio-temporal event streams, most event-based ANNs rely on frame-based representations. For example, \cite{Donahue2017LSTMCNN} proposed a hybrid CNN-RNN model where a CNN extracts spatial features that are then passed to an RNN that maintains a memory state of the eye location over time.

Another approach, inspired by computational neuroscience, is known as the spiking neural network (SNN). SNNs encode information via asynchronous spike trains, activating neurons only when needed, greatly reducing power usage~\cite{Akopyan2015TrueNorthDesignToolFlow, Ottati2023ToSpikeOrNot, Narduzzi2022OptimizingConsumptionSNN, Yin2021AccurateEfficientSNN}. Models such as \emph{Retina}~\cite{Bonazzi2024RetinaEET} and \emph{GazeSCRNN}~\cite{Groenen2025GazeSCRNN} show that SNNs can estimate gaze accurately while using less energy than standard CNNs. 

State space models (SSMs) have also been employed in order to capture long-term temporal patterns while avoiding the high computational cost of transformer architectures ~\cite{Gu2024Mamba, truong2025dual}. Bayesian variants, including \emph{recurrent state space models (RSSMs)}~\cite{Ha2018WorldModel, Hafner2020DreamerV1, Hafner2021DreamerV2, Hafner2025DreamerV3} 
model the sequence dynamics probabilistically and have been applied to eye-tracking, e.g., as in \emph{Mamba}~\cite{Wang2024MambaPupil}. In this work, we introduce and design a special variant of the SSM, further formulated in a Bayesian manner, that learns to represent the state space of the event frame observation.

\subsection{Metrics and Datasets}
\label{sec:benchmarks_metrics}

\noindent
\textbf{Metrics.} Model performance in feature extraction is commonly assessed by measuring the spatial accuracy of feature localization relative to a known ground truth. Recent work has increasingly relied on threshold-based detection rates to quantify performance. In particular, evaluations of feature-detection networks often report prediction accuracy as the cumulative proportion of samples whose estimated locations fall within a fixed pixel radius of the ground-truth pupil centroid, e.g., P$5$ denotes estimates within $5$ pixels of the labeled position~\cite{Wang2024EventBasedEyeTrackingChallenge2024, Chen2025EventBasedEyeTrackingChallenge2025}.

A key limitation of these P-metrics is that they enable meaningful comparisons only across datasets with similar image resolutions and optical imaging configurations, since differences in the eye–sensor distance, lens properties, and overall magnification directly determine the apparent size of the eye and pupil within the sensor’s field of view. Consequently, a $5$-pixel localization error is not comparable across images with different pupil sizes or resolutions; this limits the interpretability and generalizability of these metrics across studies. Here, we adopt these P-metrics in the absence of suitable alternatives, with the caveat that comparisons of P-metrics across manuscripts are only valid if evaluations were performed on similar test-sets and resolutions at evaluation time.

\noindent
\textbf{The 3ET+ Dataset.} We evaluate competing algorithms using the widely adopted \emph{3ET+} dataset~\cite{Wang2024EventBasedEyeTrackingChallenge2024, Chen2025EventBasedEyeTrackingChallenge2025}, which provides large-scale, real-world event-based eye recordings with high-quality annotations. In principle, supervised models should be trained and evaluated on datasets that capture the range of task-relevant variability expected in real-world deployment, including differences in imaging system characteristics, participant physiology and demographics, environmental illumination, task demands, and resulting gaze behaviors. 
As with other commonly used datasets, e.g., \emph{RGBE-Gaze}~\cite{Zhao2025RGBEGazeEventBasedDataset} and \emph{EV-Eye}~\cite{Zhao2023EVEyeDataset}, 3ET+ does not comprehensively sample these dimensions of variability. Specifically, it exhibits limited diversity in sensor and optical configurations, participant demographics, and recording conditions. Nevertheless, we adopt 3ET+ since it is a well-established benchmark in the event-based eye-tracking literature, supports direct comparison with prior work, and offers sufficiently large, consistently annotated data for enabling controlled and reproducible evaluation of competing methods.

\section{Methods}
\label{sec:method}

Typically, CNNs designed to accept frame-based representations of event streams exhibit limited utility when applied to event-based eye data, especially during periods of minimal eye motion when the signal-to-noise ratio or event density drops (see Section~\ref{sec:intro}). One common approach to resolve this issue is to augment the CNN with recurrent elements, as in the \emph{CB-ConvLSTM}~\cite{Chen20233ETCBConvLSTM}, \emph{MambaPupil}~\cite{Wang2024MambaPupil}, and \emph{PEPNet}~\cite{Ren2024PEPNet}. However, 
conventional RNN models often exhibit unstable performance due to their simultaneous reliance on both past and present information. To overcome this limitation, we introduce a multi-stage training pipeline for eye feature tracking. First, we employ a temporally adaptive technique that enables the system to dynamically adjust the relative weighting of current and past observations (see Section~\ref{sec:aissm}). This relative weighting is determined by the \emph{dynamic confidence network}, which dynamically adjusts this weighting on the basis of both SNR and event density (see Section~\ref{sec:conf}). Lastly, we propose a training scheme to improve the efficiency and robustness of the eye tracking model (Section~\ref{sec:horizon}).

\begin{figure*}[!t]
    \centering
    \includegraphics[width=0.9\linewidth]{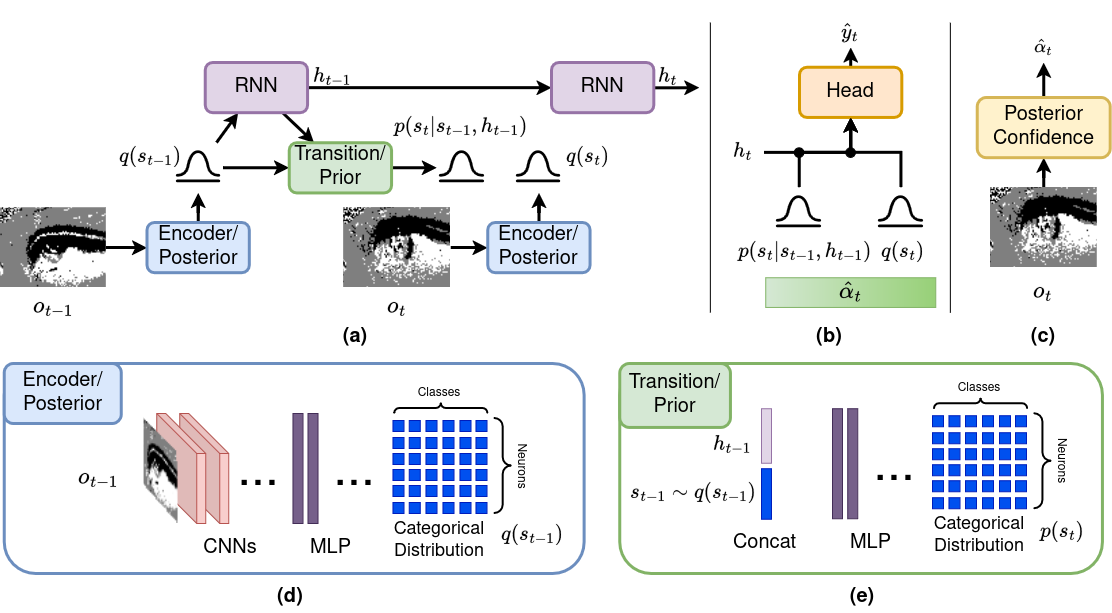}
    \caption{
    Overall architecture of our \emph{adaptive inference state space model}: 
    \textbf{(a)} The core architecture. The encoder forms a posterior distribution of the current event information $q(s_{t-1})$ while the transition model utilizes a recurrent neural network (RNN) to produce a prior distribution over the next frame $p(s_t|s_{t-1},h_{t-1})$. The RNN output ($h_{t-1}$) is also used to update the prior at the next timestep. 
    \textbf{(b)} The head is responsible for the estimation of the eye feature $\hat{y}$. Its input is an $\alpha$-weighted linear combination of the prior and posterior distributions over the latent state, i.e., it is a weighted summation of current and prior event information. 
    \textbf{(c)} The \emph{dynamic confidence network} predicts the weighting term $\alpha$ that is used in \textbf{(b)}, which is an estimate of the current information's `reliability'.  
    \textbf{(d)} The inner workings of the AISSM's encoder module, which is a combination of CNNs and a multi-layered perceptron (MLP), whose output is reshaped into a $2$D (categorical distribution) matrix that represents the present data state. 
    \textbf{(e)} The inner workings of the AISSM's transition module, which is a concatenation of the prior recurrent state ($h_{t-1}$) and the representation state ($s_{t-1}$) sampled from the previous posterior $q(s_{t-1})$. This input is passed through an MLP, forming the prior (categorical) distribution over past data $p(s_t)$.
    }
    \label{fig:aissm}
\end{figure*}

\subsection{The Adaptive Inference State Space Model (AISSM)}
\label{sec:aissm}

To address the problems inherent to using RNNs, where performance degrades when event density/SNR changes, we introduce a model that is designed to dynamically weight the contribution of past and present information. 
Since there is only one event frame in the present, whereas there are multiple event frames in the past, our model combines past and present information in the context of a latent code. Specifically, we represent this possible latent information using a discrete, categorical distribution~\cite{Hafner2021DreamerV2, Hafner2025DreamerV3, nguyen2025sr} instead of a single latent vector in contrast to what is typically done in standard variational inference~\cite{Kingma2014VAE, Kalman1960Filter}, active inference~\cite{Friston2003InferenceBrain, Friston2010FEP, Friston2013LifeAsWeKnowIt, Ororbia2023ActivePredictiveCoding, Yang2023Neural}, and representation/model-based-reinforcement learning~\cite{Ha2018WorldModel, Hafner2019Planet, Hafner2020DreamerV1, Hansen2022TDMPC, Hansen2024TDMPC2} schemes. We then compute the downstream latent representation using the dynamically-weighted linear combination between the past and the present information. We refer to this modeling approach as the \emph{adaptive inference state space model (AISSM)}.


\noindent
\textbf{Model Formulation.} Different parts of our proposed AISSM are constructed utilizing ANNs and can be expressed in the following manner:
\begin{equation}
\begin{aligned}
    \text{Encoder Posterior: } &q_\theta (s_t | o_t); \\
    \text{Recurrent Dynamics: } &f_\omega(h_t | h_{t-1}, s_{t-1}) \\
    \text{Transition Prior: } &p_\nu(s_t | h_t); \\
    \text{Confidence: } &f_\lambda(\hat{\alpha}_t | o_t); \\
    \text{Task Head: } &f_\psi(\hat{y}_t | \bar{s}_t). 
\end{aligned}
\end{equation}
In this formulation, $\theta, \omega, \nu, \lambda, \psi$ are the ANN parameters of our model's corresponding encoder, recurrent model, prior, confidence, and task head, respectively. 
$o_t \in \mathcal{O}^{H \times W \times X}$ is the event frame observation at time $t$ with height $H$ and width $W$ and $X$ is the number of channel(s), depending on the event representation method. Similarly, $s_t \in \mathbb{R}^{d_S} $ is the corresponding state representation with dimension $d_S$. Next, $h_t \in \mathbb{R}^{d_R}$ is the recurrent state with dimension $d_R$. The (dynamic) confidence network $f_\lambda$ maps the observation space to a single term $\hat{\alpha}_t \in \mathbb{R}^1, \hat{\alpha}_t \in [0, 1]$, which determines how reliable the signal is within the most current observation (event frame). $\alpha_t$ is then used in a linear combination process so as to produce a dynamically weighted state representation $\bar{s}_t$, i.e., $L(s^q_t, s^p_t) = \bar{s}_t = \hat{\alpha}_t \times s^q_t + (1 - \hat{\alpha}_t) \times s^p_t; \ s^p_t \sim p_\nu(s_t | h_t),\ s^q_t \sim q_\theta(s_t | o_t)$. 
Finally, the weighted state representation $\bar{s}_t$ is used as input to the task prediction head $f_\psi$ in order to produce the final prediction of eye feature $\hat{y}_t \in \mathbb{R}$. Note that the dimension of $\hat{y}$ depends on the dataset's label expectation; the 3ET+ dataset~\cite{Chen20233ETCBConvLSTM, Wang2024EventBasedEyeTrackingChallenge2024, Chen2025EventBasedEyeTrackingChallenge2025} uses pupil centroid ($x$ and $y$ pixel coordinate) as the label output, so $\hat{y}_t \in \mathbb{R}^2$. Model details are further described in Figure~\ref{fig:aissm}.


\noindent
\textbf{Model Training.} Unlike variational inference approaches~\cite{Kingma2014VAE, Karl2017VariationalBayesFilters}, AISSM does not aim to collapse the posterior and prior distributions to update beliefs based on observation-induced \emph{surprisal}. Continually collapsing the posterior and prior distributions would cause the posterior to become fully informed by past states, effectively reducing the AISSM to a conventional RNN. 
In contrast, we treat the posterior distribution as purely informative of the present and the prior distribution as purely informative of the past. Therefore, no complexity objective/penalty~\cite{Friston2016AIFandLearning}, such as the KL divergence between posterior and prior, was required. Finally, the downstream task head can learn to predict the eye features using any distance loss. In our work, we employ the Huber loss~\cite{Huber1964HuberLoss}, denoted as $\mathcal{U}$, so as to ensure a more stable learning trajectory. The AISSM learning objective can then be expressed as follows:
\begin{equation}
\begin{aligned}
    \underset{\theta, \omega, \nu, \psi}{\arg \min} \ \ \mathcal{L}_t (\theta, \omega, \nu, \psi) = \  &\mathcal{U} \bigl[ y_t, f_\psi (L (s^q_t, s^p_t) ) \bigr]; \\
    s^p_t \sim p_\nu(s_t | f_\omega (h_t | h_{t-1}, &s_{t-1}) ),\ s^q_t \sim q_\theta(s_t | o_t)\\
     \text{where, Huber loss: } &\mathcal{U} \bigl[ y, \hat{y} \bigr] = \\
     0.5 \min(|\hat{y} - y|, \delta)^2 + &\delta \bigl( |\hat{y} - y| - \min(|\hat{y} - y|, \delta) \bigr). 
\end{aligned}
\end{equation}
Note that $\delta$ is the threshold where the Huber loss transitions from a quadratic objective, e.g., the mean squared error, to a linear objective, e.g., the mean absolute error. This term controls the model's outlier sensitivity as well as clips the gradients, further stabilizing our AISSM's training objective; in our experiments, we set $\delta = 1.0$. Finally, since the sampling process from a categorical distribution is non-differentiable, we employ the straight-through gradient technique to facilitate the training of the AISSM. Specifically, one can implement the sampling of the state from the categorical as follows:
\begin{equation}
\begin{aligned}
    s \sim q(s) \Leftrightarrow s = \text{sg} \bigl( \text{sample} (\mathbf{z}) \bigr) + \Bigl( \sigma(\mathbf{z}) - \text{sg}\bigl( \sigma(\mathbf{z}) \bigr) \Bigr)  
\end{aligned}
\end{equation}
where $s$ is the resulting discrete state represented as a one-hot vector, $q(s)$ is a categorical distribution, and $\mathbf{z}$ is the parameter of that distribution. With $\text{sample}(\cdot)$ and $\text{sg}(\cdot)$ denoting the normal categorical sampling and the stop-gradient functions, respectively, the ANN forward pass uses a hard sample and the backward pass uses the gradient of the softmax function $\sigma(\mathbf{z})$.

\subsection{The Dynamic Confidence Network}
\label{sec:conf}

\textbf{Model Formulation.} To dynamically infer the confidence of the signal contained within the current observation to be used in a final downstream task head, we require another module to examine and score the actual observation, i.e., $f_\lambda(\hat{\alpha}_t | o_t)$. We define the confidence score of the frame based on the SNR and the event density and train an ANN to predict the total confidence score computed using these two terms. 
Formally, we first define a region of interest (ROI) $r_t$, which has height $h$ and width $w$, by forming a region around the label of each event frame. Specifically, $r_t$ has its top-left pixel coordinate set to $(y^i_t - \frac{h}{2}, y^j_t - \frac{w}{2})$ and the bottom-right coordinate set to $(y^i_t + \frac{h}{2}, y^j_t + \frac{w}{2})$, where $i, j$ are the $i$-th row and $j$-th column in the event frame, respectively. Furthermore, $h$ and $w$ are based on the ratio between the eye and the frame. In our experiments with training resolution of $160 \times 120$, we select $h=40$ and $w=70$. 
The label confidence score $\alpha_t$ can then be computed in the following manner:
\begin{equation}
\begin{aligned}
    \text{SNR}_t = \sigma \left( \frac{\sum_{i,j} e^{i,j}_t \in r_t}{ \sum_{i,j} e^{i,j}_t \notin r_t } \right); \ \ 
    \text{ED}_t = \left. \frac{\sum_{i,j} e_t^{i,j} \in r_t}{h \times w \times \tau} \middle| \right|_{0}^{1}; \ \alpha_t = \beta \ \text{SNR}_t + (1 - \beta) \ \text{ED}_t
\end{aligned}
\end{equation}
where the signal-to-noise ratio $\text{SNR}_t$ is first computed by calculating the ratio between the sum of events inside the ROI $r_t$ and the sum of events outside of $r_t$. This quantity is then normalized to the range $[0, 1]$ using a sigmoid squashing function. 
Next, the event density $\text{ED}_t$ is defined by calculating the ratio between the total number of events in $r_t$ and the number of pixels that $r_t$ contains ($h \times w$). Since the event stream is sparse, even within an event frame, we normalize the number of pixels in $r_t$ down to a threshold $\tau = 0.1$, forming the denominator $h \times w \times \tau$. 
$\text{ED}_t$ is then clipped to remain between $0$ and $1$ and, finally, the label confidence score $\alpha_t$ is computed by calculating the weighted average of the $\text{SNR}_t$ and $\text{ED}_t$ via the term $\beta$. We found that smaller beta values, e.g., $\beta = 0.1$, helped to improve the training and inference stability of the model. Given the labeled confidence score, we can train the \emph{dynamic confidence network} using any distance objective. Similar to training the AISSM downstream task head, we optimize this module using a Huber loss~\cite{Huber1964HuberLoss}:
\begin{equation}
\begin{aligned}
    \underset{\lambda}{\arg \min} \mathcal{L}_t (\lambda) = \mathcal{U} \bigl[ \alpha_t, f_\lambda(\hat{\alpha}_t | o_t) \bigr].
\end{aligned}
\end{equation} 
Note that the ROI and the label confidence $\alpha_t$ depends on the accuracy of the eye feature labels in the dataset. An imperfect label can cause the ROI to shift away from the eye region, thus reducing the accuracy of the dynamic confidence network. In this work, we assume that the labeled eye features are accurate.

\subsection{Long-Horizon Training Technique}
\label{sec:horizon}

When training recurrent/temporal algorithms, one encounters a problem known as \emph{stateless training}~\cite{Schafer2008LearningLongTermDependenciesRNN, Johnston2025RevisitingLongTermDependenciesRNN, Trinh2018LongerTermDependenciesRNNAuxiliaryLosses}. 
Specifically, when the recurrent state of the RNN is re-initialized at every training batch, as is normally done when training RNNs, temporal information is also reset and therefore lost. One can resolve this issue by storing the model state and only resetting it at each epoch's start. However, one epoch might contain multiple independent gaze behaviors along unique trajectories. To prevent interference among these largely interdependent behaviors, we introduce a training technique for typical sequential, RNN-based models. Specifically, we attach the model state, e.g., posterior and prior distributions, and the recurrent state of the internal RNN directly to each training item and store them within the dataset. Upon training each instance, we retrieve a new state for that instance, and then update it to the corresponding item within the dataset. Our method differs from truncated back-propagation through time by utilizing a replay buffer, i.e., typically used in reinforcement learning~\cite{Haarnoja2018SAC}, in order to achieve uniform sampling and update the model state to the correct dataset index after training every batch.

\section{Experimental Results}
\label{sec:results}

We design our experiments to address and fulfill the following key research objectives:
\begin{enumerate}[noitemsep, topsep=2pt]
    \item show that our model outperforms state-of-the-art baselines with respect to the accuracy of eye feature (e.g., pupil centroid) estimation; 
    \item compare the behavior of our AISSM to other baseline methods in the context of sudden change in event density / eye movement behavior; and, 
    \item demonstrate the performance improvement of our proposed long-horizon training mechanism.
\end{enumerate}

\textbf{Simulation Setup.} We compare our method with recent strong baselines, including \emph{CB-ConvLSTM} \cite{Chen20233ETCBConvLSTM}, a CNN with a bidirectional gated recurrent unit (\emph{CNN-BiGRU}) as used in~\cite{Schuster1997BidirectionalRNN, Chung2014GRU}, and \emph{MambaPupil}~\cite{Wang2024MambaPupil} (the winner of CVPR's event-based eye tracking challenge~\cite{Wang2024EventBasedEyeTrackingChallenge2024}). In addition, we train/simulate general baseline algorithms: 
1) a convolution neural network (\emph{CNN})~\cite{Lecun1995CNN}; 
2) a CNN equipped with a gated recurrent unit~\cite{Chung2014GRU} (\emph{CNN-GRU}); 
and 
3) CNN integrated with a Mamba~\cite{Gu2024Mamba, Dao2024Mamba2} module (\emph{CNN-Mamba}). 
We compare each method with respect to their accuracy performance, i.e. prediction success rate within $5$ pixels (P$5$), $10$ pixels (P$10$), and $15$ pixels (P$15$), and the distance error which measures how far away the predicted pupil is from the labeled pupil. Note that the distance metrics are normalized to the range $[0, \sqrt{2}]$, i.e., since $x$ and $y$ coordinates are normalized the range becomes $[0, 1]$. For fair comparison, all models are constrained to $\approx 500,000$ (synaptic) parameters. In accordance with neural scaling laws~\cite{Kaplan2020NeuralScalingLaws}, we assume that the comparison across different model with higher number of parameters would yield similar ratio. 
Finally, the 3ET+ dataset~\cite{Wang2024EventBasedEyeTrackingChallenge2024, Chen2025EventBasedEyeTrackingChallenge2025} is used as the main benchmark for comparison. 

To preprocess the data, we follow the methodologies of \emph{MambaPupil} \cite{Wang2024MambaPupil}, \emph{CB-ConvLSTM} \cite{Chen20233ETCBConvLSTM}, \emph{CETM} \cite{Wang2024EventBasedEyeTrackingSurvey}, \emph{TENNs} \cite{Pei2024LightweightSpatioTemporalNetwork}, and \emph{TDTracker} \cite{Ren2025TDTracker}; event frames/labels are transformed to match network inputs and training efficiency is improved by lowering input resolution to $160 \times 120$. To construct an event representation, we use \emph{binarep} \cite{Barchid2022BinaRepEventFrames} which yields higher accuracy in the \emph{CB-ConvLSTM} baseline \cite{Chen20233ETCBConvLSTM}. After training each model instance, we compute the mean and standard deviation over $5$ runs on the validation set. Our final test-time metrics are evaluated at $320 \times 320$ resolution in order to closely match the setup of prior 
pipelines \cite{Scheerlinck2019CEDDataset, Angelopoulos2021EventEye10000Hz}.

\begin{table*}[t]
\centering
\caption{Test/validation accuracy, P$5$, P$10$, P$15$, and normalized predicted error distance metrics for each model in the 3ET+ dataset at a resolution of $320 \times 320$ pixels. \textuparrow  Higher mean is better. \textdownarrow Lower mean is better. Note that cells highlighted in blue/cyan denote the best performance. All models were implemented under this work's experimental parameter constraints in order to ensure fair comparison.}
\begin{tabular}{p{2.68cm}|cccccccc}
	\textbf{Model/Metrics} & \textbf{P$5$ \textuparrow} & \textbf{P$10$ \textuparrow} & \textbf{P$15$ \textuparrow} & \textbf{Distance \textdownarrow} \\\hline\hline
	CNN & $17.67 \pm 10.88$ & $41.82 \pm 20.04$ & $60.17 \pm 20.48$ & $0.03 \pm 0.01$ \\
	CNN-GRU & $25.33 \pm 14.28$ & $48.62 \pm 26.66$ & $62.44 \pm 28.47$ & $0.03 \pm 0.01$ \\
	CNN-Mamba & $16.02 \pm 14.38$ & $45.43 \pm 39.38$ & $55.61 \pm 41.01$ & $0.03 \pm 0.01$ \\	\hline
	CNN-BiGRU & $21.63 \pm 21.48$ & $41.95 \pm 41.47$ & $49.33 \pm 48.00$ & $0.03 \pm 0.01$ \\
	CB-ConvLSTM & $29.86 \pm 15.30$ & $57.40 \pm 21.25$ & $71.94 \pm 18.21$ & $0.03 \pm 0.01$ \\
	MambaPupil & $11.35 \pm 11.33$ & $32.84 \pm 32.37$ & $48.22 \pm 45.95$ & $0.13 \pm 0.06$ \\	\hline
	\textbf{AISSM (ours)} & \cellcolor{LightCyan}$46.85 \pm 21.49$ & \cellcolor{LightCyan}$72.38 \pm 8.13$ & \cellcolor{LightCyan}$85.41 \pm 0.79$ & \cellcolor{LightCyan}$0.01 \pm 0.01$ \\
\end{tabular}
\label{tab:stats}
\end{table*}

\noindent
\textbf{Experimental Results.} Overall, with the constraint of only $\approx 500,000$ parameters in total, our AISSM model achieves the best performance as compared to other baselines with respect to P$5$, P$10$, P$15$, as well as normalized distance (with better reported mean and relatively lower standard deviation statistics); see Table~\ref{tab:stats}. In addition, AISSM exhibits a significantly lower standard deviation than the baselines. We attribute this performance fluctuation (of baseline models) to sudden changes in eye behavior / event density. This effect is further intensified in the evaluation phase where the testing resolution is higher.
This empirical result demonstrates that our AISSM is capable of dynamically weighting the information coming from the past and the most current observation, facilitating effective eye feature estimation.

\begin{minipage}[t]{0.48\linewidth}
    \begin{figure}[H]
        \centering
        \includegraphics[width=\linewidth]{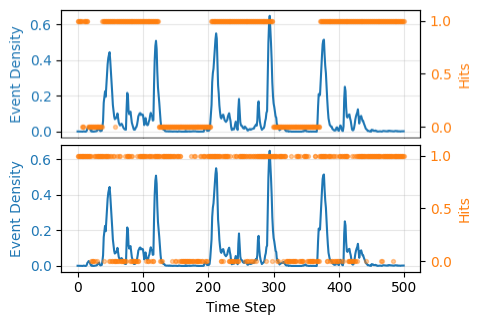}
        \caption{The relationship between event density (blue) and successful prediction (hit) within the range of $10$ pixels (orange). The $x$-axis represents the time-steps within an event frame. Top depicts the CNN-GRU; bottom depicts the AISSM.}
        \label{fig:edhits}
    \end{figure}
\end{minipage}
\hfill
\begin{minipage}[t]{0.48\linewidth}
    \begin{figure}[H]
        \centering
        \includegraphics[width=\linewidth]{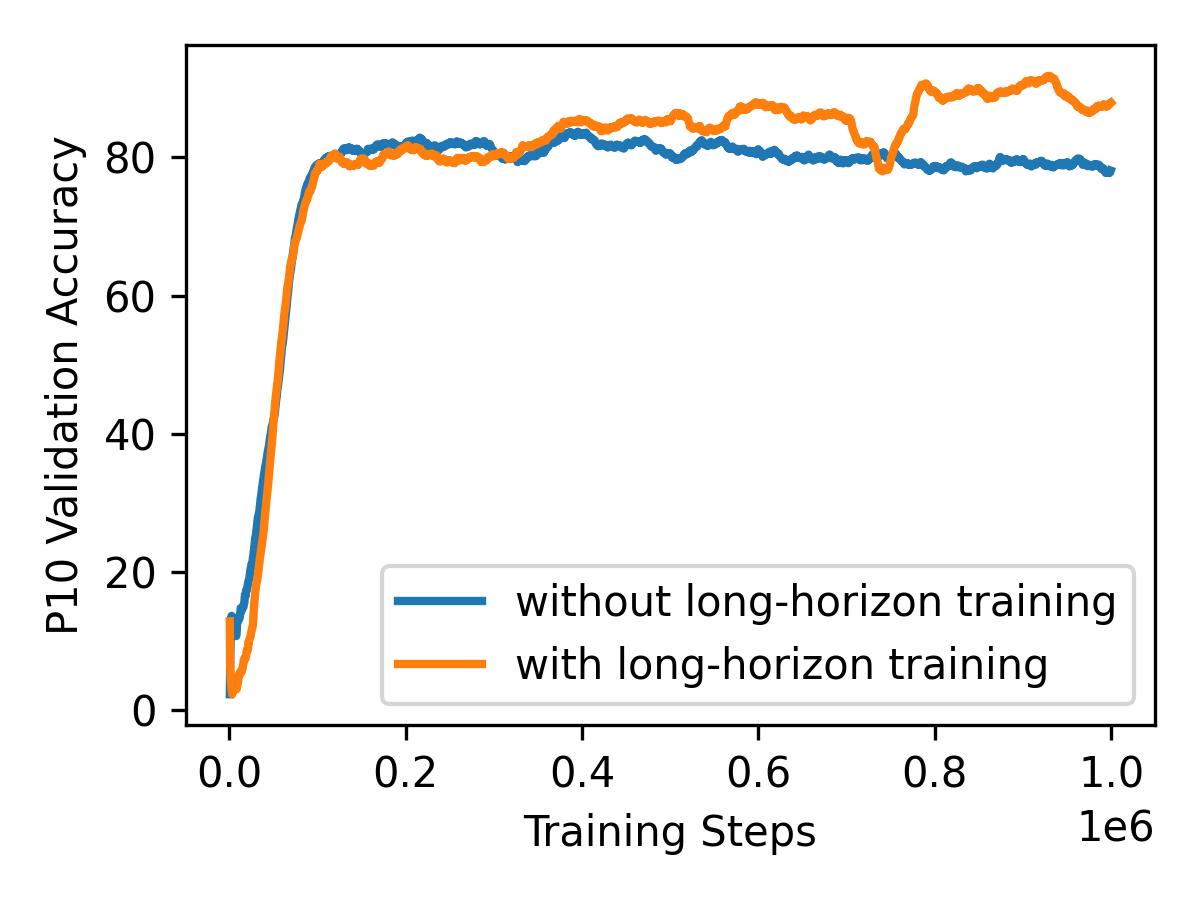}
        \caption{Our AISSM's validation accuracy when training with (orange) and without (blue) our proposed long-horizon training technique.}
        \label{fig:longeff}
    \end{figure}
\end{minipage}

Furthermore, Figure~\ref{fig:edhits} illustrates the utility of being able to change the relative weighting on history and current information on the basis of the quality of current information, as indicated by an estimate of event density. The interaction between current and recent information is most apparent at transitions between gaze behaviors; for instance, when the eye suddenly transitions from saccade to fixation. Inspection of these saccade-to-fixation transitions led us to the speculate that RNN-centric models such as the CNN-GRU rely heavily on the current observation, resulting in rapid performance deterioration as event density drops during fixation. In contrast, as event density drops, our AISSM dynamically increases its weight on past information in response to low posterior confidence; this results in a more informative input to the downstream task head. As a result, the number of ``hits'' (<P$10$) produced by the AISSM exceeds that of the CNN-GRU model.

Lastly, we measure the performance of our model trained with and without the novel long-horizon technique. Figure~\ref{fig:longeff} shows the stabilizing effect that is provided by our proposed training method. Specifically, with the storage and reuse of model states, the accuracy of our trained model stabilizes over time as the training steps become larger; this usefully avoids the slight divergence and degradation in accuracy that happens later on when the model is trained without this technique.

\section{Conclusion}
\label{sec:conclusion}

We introduced the \emph{adaptive inference state space model (AISSM)} for enhancing eye feature estimation from event data streams. This Bayesian model learns to dynamically weight the information between past and present event frames through an inferred confidence scoring scheme. A novel \emph{dynamic confidence network} learns to estimate the labeled confidence at training time. This labeled confidence is calculated based on the region of interest inferred from the actual position of the labeled eye feature (e.g., the pupil centroid), event density, and signal-to-noise ratio. Empirically, we demonstrate that our proposed AISSM system outperforms other state-of-the-art eye-tracking model baselines. Project-related source code has been made publicly available at \hyperlink{https://github.com/PerForm-Lab-RIT/event-based-eye-tracking-pipeline}{https://github.com/PerForm-Lab-RIT/event-based-eye-tracking-pipeline}. 

\noindent
\textbf{Limitations and Future Work.} Although our method can outperform other state-of-the-art baselines at a similar number of parameters, the training/inference speed has not been taken into account and could potentially yield higher latency due to the existence of two networks (the AISSM and \emph{dynamic confidence network}). Furthermore, our experiments are conducted on a single dataset and potentially raise a question of generalizability. Future work might resolve this potential problem by testing on multiple different datasets such as EV-Eye~\cite{Zhao2023EVEyeDataset} and RGBE-Gaze~\cite{Zhao2025RGBEGazeEventBasedDataset} and investigating different model optimization methods such as using observation uncertainty-aware confidence~\cite{Friston2015Epistemic, Friston2017GraphicalBrain} instead of a separate network. One could also propose a different representation method, i.e., using disentanglement~\cite{Wang2023DisentangledRepr, Higgins2018DisentangledRepr}, instead of employing a categorical distribution as we do, since training and sampling might be temporally/computationally expensive.

\section*{Acknowledgments}
We would like to thank Meta Reality Labs for providing invaluable feedback and supporting this work.\footnote{Authors Alexander Fix and Ryan J. Suess at Meta were involved as advisors on this research, which was conducted at RIT.} This material is based upon work supported by the National Science Foundation under Award No. DGE-2125362. Any opinions, findings, and conclusions or recommendations expressed in this material are those of the author(s) and do not necessarily reflect the views of the National Science Foundation.

\bibliographystyle{ACM-Reference-Format}
\bibliography{main}

\end{document}